\documentclass[letterpaper]{article} 
\usepackage[]{aaai23}  
\usepackage{times}  
\usepackage{helvet}  
\usepackage{courier}  
\usepackage[hyphens]{url}  
\usepackage{graphicx} 
\usepackage{natbib}
\usepackage{caption}
\usepackage{newfloat}
\usepackage{listings}
\DeclareCaptionStyle{ruled}{labelfont=normalfont,labelsep=colon,strut=off} 

\usepackage{mathtools}
\usepackage{amsmath}
\usepackage{pgfplots}
\usepackage{algpseudocode}
\usepackage{amssymb}
\usepackage{textcomp}
\usepackage{booktabs}
\usepackage{placeins}
\usepackage{subfig}

\pgfplotsset{compat=newest}

\newcommand{\alg}{PVD-B}
\newcommand{\perf}{$55\%$}

\newcommand{\argmax}[1]{\underset{#1}{\operatorname{arg} \operatorname{max}}\;}

\title{Dynamic Pricing with Volume Discounts in Online Settings}
\author{
    Marco Mussi\textsuperscript{\rm 1}\equalcontrib, Gianmarco Genalti\textsuperscript{\rm 1}\equalcontrib, Alessandro Nuara\textsuperscript{\rm 2},\\
    Francesco Trov\'{o}\textsuperscript{\rm 1}, Marcello Restelli\textsuperscript{\rm 1}, { \normalfont and} Nicola Gatti\textsuperscript{\rm 1}\\
}
\affiliations{
    \textsuperscript{\rm 1}Politecnico di Milano, Milan, Italy\\
    \textsuperscript{\rm 2}ML cube, Milan, Italy\\
    \{marco.mussi, gianmarco.genalti, francesco1.trovo, marcello.restelli, nicola.gatti\}@polimi.it \\
    alessandro.nuara@mlcube.com
}

\begin{document}

\maketitle

\begin{abstract}
    According to the main international reports, more pervasive industrial and business-process automation, thanks to machine learning and advanced analytic tools, will unlock more than $14$ trillion USD worldwide annually by $2030$. In the specific case of pricing problems---which constitute the class of problems we investigate in this paper---, the estimated unlocked value will be about $0.5$ trillion USD per year.
    In particular, this paper focuses on pricing in e-commerce when the objective function is profit maximization and only transaction data are available.
    This setting is one of the most common in real-world applications.     
    Our work aims to find a pricing strategy that allows defining optimal prices at different volume thresholds to serve different classes of users.
    Furthermore, we face the major challenge, common in real-world settings, of  dealing with limited data available. 
    We design a two-phase \emph{online learning} algorithm, namely \alg{}, capable of exploiting the data incrementally in an online fashion. 
    The algorithm first estimates the demand curve and retrieves the optimal average price, and subsequently it offers discounts to differentiate the prices for each volume threshold. 
    We ran a real-world $4$-month-long \textsf{A}/\textsf{B} testing experiment in collaboration with an Italian e-commerce company, in which our algorithm \alg{}---corresponding to \textsf{A} configuration---has been compared with human pricing specialists---corresponding to \textsf{B} configuration. 
    At the end of the experiment, our algorithm produced a total turnover of about $300$ KEuros, outperforming the \textsf{B} configuration performance by about \perf{}. The Italian company we collaborated with decided to adopt our algorithm for more than 1,200 products since January 2022.
\end{abstract}

\section{Introduction} \label{sec:introduction}

Most international economic forecasts agree that nearly $50\%$ of the annual value  unlocked by the adoption of Artificial Intelligence (AI) from $2030$ on will be in marketing\&sales~\cite{chui2018notes}. Examples of 
activities in which AI tools can play a central role for marketing\&sales include attracting and acquiring 
new customers, suggesting and recommending products, and optimizing customers' retention and loyalty. In particular, AI can effectively automate these processes so as to increase their efficiency dramatically.

This paper focuses on pricing for e-commerce when, as it is usual, the objective is profit maximization and only transaction data are available. In particular, we focus on settings in which an e-commerce website sells goods other than luxury, Veblen, and Giffen. Thus, we can assume, without loss of generality, that the demand curve is monotonically decreasing in price. Furthermore, we assume that the e-commerce website works with different classes of customers both in B2B and B2C scenarios. However, at the stage the price of the product is chosen and displayed to a user, the seller does not know whether the user comes from the former or latter scenario. Usually, volume discount is used to deal with multiple classes of users when it is not possible to distinguish the classes at the price formation stage. In particular, the rationale is to propose different \emph{prices} for different \emph{volume 
thresholds} thanks to the introduction of discounts. This approach allows showing the same thresholds and prices to all incoming users and, at the same time, it introduces price discrimination to provide a 
different pricing strategy for different classes of users.
To the best of our knowledge, even if the problem of learning the price that maximizes the seller's revenue has been extensively 
studied in the economic~\cite{klenow2010microeconomic}, game theory~\cite{kopalle2010game} and learning~\cite{den2015dynamic} 
fields, no dynamic pricing algorithm in the literature deals with volume discounts in a data-driven way.

\subsubsection{Original Contribution}

In this work, we design an online-learning pricing algorithm, namely the \textit{Pricing with Volume Discounts Bandit} (\alg) algorithm.
We face the problem of assigning different prices to different volume thresholds using transaction data (coming from historical purchases and during the operational life of the e-commerce website). Given the complex dynamics of the problem, we decompose the algorithm into two phases: an optimal average price estimation and, based on the above estimation, a price adaptation method 
to provide different prices for the given volume discount thresholds. The adoption of tools from online learning guarantees convergence to optimal prices.

In collaboration with an Italian e-commerce website, we ran a real-world $4$-month-long \textsf{A}/\textsf{B} testing experiment over a set of $\approx 300$ products, in which our 
algorithm \alg{}---corresponding to \textsf{A} configuration---has been compared with human pricing specialists---corresponding 
to \textsf{B} configuration. At the beginning of the test, the available data concerned the purchases occurred in the previous $2$ years. 
At the end of the experiment, 
the total turnover of \textsf{A} configuration was more than $300$~KEuro and
our algorithm \alg{} performed better than the \textsf{B} configuration in terms of the objective 
function (\emph{i.e.}, total profit) for about \perf{}. The company we collaborated with decided to adopt our algorithm for more than 1,200 products since January 2022.

\section{Related Works} \label{sec:relatedworks}

A comprehensive analysis of the dynamic pricing literature is provided in~\citet{narahari2005dynamic,bertsimas2006dynamic,den2015dynamic}.
In particular, Multi-Armed Bandits (MAB) techniques have been extensively employed for dynamic pricing when the available information concerned the interactions between the e-commerce website and customers.

\citet{rothschild1974two} presents one of the seminal works on the adoption of MAB algorithms for dynamic pricing. This algorithm has been 
subsequently extended in several directions to capture the characteristics of different pricing settings. \citet{kleinberg2003value} study the problem of dealing with continuous-demand functions and proposes a discretization of the price values to provide theoretical guarantees on the regret of the algorithm. This approach suffers from the drawback that the reward is assumed to have a unique maximum in the price. 
Such an assumption is hard to be verified in practice.
Instead, \citet{trovo2015multi,trovo2018improving} relaxed this assumption, assuming that the demand function is monotonically decreasing and exploiting this assumption in the learning algorithm to provide uncertainty bounds tighter than those of classical frequentist MAB algorithms. However, the model formulation explicitly imposes neither monotonicity nor weak monotonicity on the estimated demand functions, so decisions that violate business logic can be allowed during the learning process.
The authors show how the monotonicity assumption does not improve the asymptotic bound of regret provided by the MAB theory. 
On the other hand, exploiting monotonicity allows for an empirical improvement in performance~\citep{mussi2022pricing}.
The same argument also holds for the work proposed by~\citet{misra2019dynamic}, where the monotonicity property of the demand function is used to ensure faster convergence. However, monotonicity is not forced as a model-specific feature.
\citet{besbes2015surprising} show that linear models are a suitable and efficient tool for modeling a demand function.
In their work, downward monotonicity is forced on a model-wise level, but it is only analyzed in a stationary environment. Other works that adopt a parametric formulation of the demand function are by~\citet{besbes2009dynamic,broder2012dynamic}. These works assume stationary customer behavior.
\citet{bauer2018optimal,cope2007bayesian} are two of the main works on Bayesian inference applied to dynamic pricing. 
They both fail to impose monotonic constraints on the model. Interestingly, \citet{bauer2018optimal} take into account non-stationary features (\emph{e.g.}, competitors' prices). \citet{araman2009dynamic} use a Bayesian approach to dynamic pricing using a prior belief on the parameters to capture market-related information and force the model to be monotonic.
\citet{wang2021multimodal} investigates non-parametric models for demand function estimation. In this case, the authors assume that the demand function is smooth.
Finally, \citet{nambiar2019dynamic} propose a model to deal with both the non-stationarity data and the model misspecification. 
However, the required contextual knowledge at a product-wise level is not usually available in practice.

To the best of our knowledge, the literature lacks a data-driven methodology for finding an optimal volume discount pricing 
schedule to maximize retailers' profits and revenues in a B2C environment.
The works from~\citet{hilmola2021quantity,rubin2003generalized} focus on the \emph{Economic Order Quantity} (EOQ) model that requires
demand size over an annual budget and stock size.
\citet{sadrian1992business} relax the EOQ hypothesis and provide a rational and straightforward pricing strategy that forces a lower bound on the company's expected profit by calculating volume thresholds and corresponding discounts afterward. The authors show the importance of volume discounts when increasing higher-priced products sales.

\section{Problem Formulation}
\label{sec:problemformulation}

We study the scenario in which an e-commerce website sells non-perishable products with unlimited availability. 
The assumption of independence among the products allows us to focus singularly on every product. 
The extension to the case with a set of products is straightforward.

\begin{figure}[t!]
    \centering
    \resizebox{0.8\columnwidth}{!}{\includegraphics{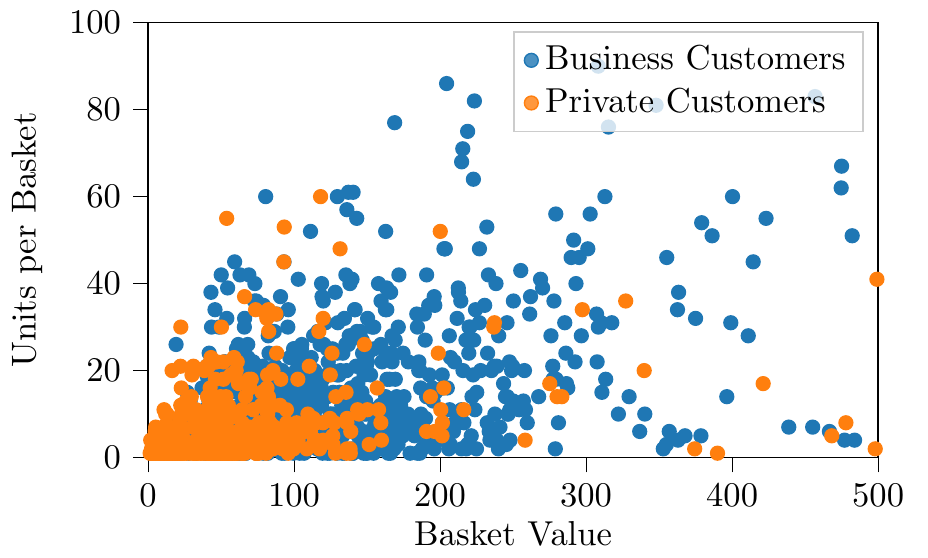}}
    \caption{Units per basket and basket values for different classes of users.}
    \label{fig:business_vs_private}
\end{figure}

Commonly, the behavior of the users purchasing items from the e-commerce website is fragmented into multiple classes.
For instance, Figure~\ref{fig:business_vs_private} provides the shopping baskets cardinality (in terms of units and economic value) for the e-commerce under analysis in this work, distinct classes of users: privates and businesses. The figure highlights how the privates purchase smaller amounts of products while the business is transacting with larger amounts. 
This fact suggests that an optimal seller strategy may include different pricing for the two. However, the user classes are not disclosed until payment is made, and, therefore, a pricing scheme that explicitly uses such a feature is not a viable option. 
In this work, we circumvent the issue of lack of customer information by using a \emph{discount threshold scheme} that differentiates the per-unit price of the items by using the number of items purchased in a single transaction as a proxy to distinguish the two classes.\footnote{
Notice that for the sake of presentation, the example presented two classes, but multiple ($> 2$) behaviors may exist, requiring multiple prices and thresholds.}\footnote{We remark that all the different prices (and related quantity thresholds) are displayed to any customer visiting the product web page.}
In addition, further benefits of such a pricing scheme have been shown in the economic literature, \emph{e.g.}, in~\citet{monahan1984quantity}. 
Indeed, this model has been shown to anticipate buyer behavior and increase average order size, allowing the retailer to access the supplier's rebate on large restocks, reduce processing costs, and anticipate cash flows through the fiscal year.

\begin{figure}[t!]
    \centering
    \resizebox{0.8\columnwidth}{!}{\includegraphics[]{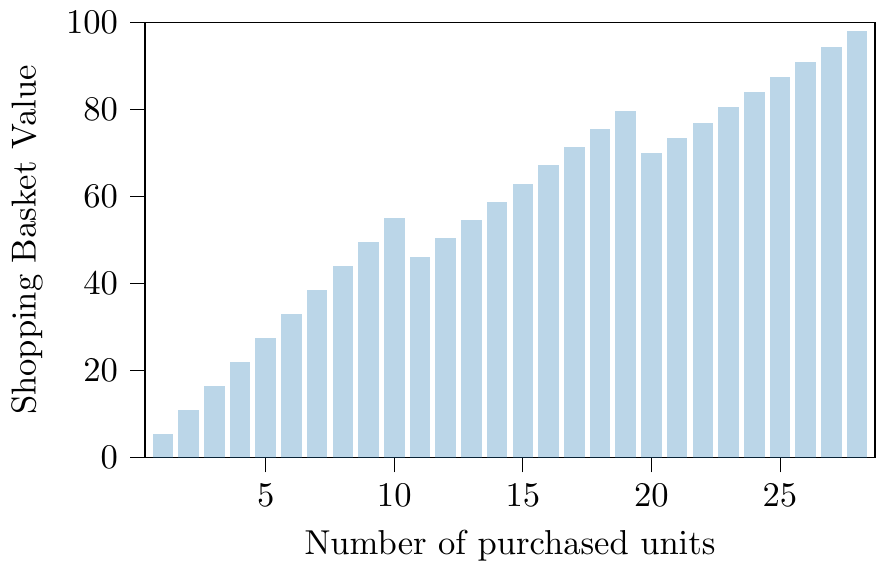}}
    \caption{Effect on the basket value given a volume discounts scheme.}
    \label{fig:vol_disc_example}
\end{figure}

Formally, for a given time $t$, let us define a vector of volume thresholds 
$\boldsymbol\omega_t := [\omega_{1t}, \ldots, \omega_{\eta t}] \in \mathbb{N}^{\eta}$, with $\omega_{it} > \omega_{ht}$, for each $i > h$ and $\omega_{1t} = 1$.
The corresponding price vector will be $\boldsymbol{p}_t := [p_{1t}, \ldots, p_{\eta t}] \in \mathcal{P}^{\eta}$, with $p_{1t} > \ldots > p_{\eta t} > 0$. $\mathcal{P}$ is the set of feasible prices, and $\eta \in \mathbb{N}$ is the number of thresholds.
The $i$-th element $p_{it}$ of $\boldsymbol{p}_t$ denotes the price proposed for each product when a user wishes to purchase 
a number of them in $\{\omega_i, \ldots, \omega_{i+1}-1\}$.\footnote{
For instance, if $\boldsymbol\omega_t = [1, 3, 5]$ 
and $\boldsymbol p_t = [6, 5, 4]$, a customer purchasing $2$ units will pay $2 \cdot 6 = 12$, and a customer purchasing 
$4$ units will pay them $4 \cdot 5 = 20$.}
Figure~\ref{fig:vol_disc_example} exemplifies the mechanism of the thresholded volume discounts when $\eta = 3$.
The seller's objective is to maximize the \emph{per-round profit}, defined as:
\begin{equation}
	R_t(\boldsymbol{p}_t, \boldsymbol{\omega}_t) := \sum_{i=1}^\eta (p_{it} - c) \cdot v_{i}(\boldsymbol{p}_t, \boldsymbol{\omega}_t, t),
\label{eq:inst_reward}
\end{equation}
where $c \in \mathbb{R}^+$ is the unit cost of an item (assumed constant) and $v_i(\boldsymbol{p}_t, \boldsymbol{\omega}_t, t) \in \mathbb{N}$ 
is the number of items sold at time $t$ and at price $p_{it}$, when the purchase consists of a number of item in $\{\omega_i, \ldots, \omega_{i+1}-1\}$ and the overall seller strategy consists of prices $\boldsymbol{p}_t$ and thresholds $\boldsymbol{\omega}_t$.\footnote{
Let us remark that this formulation can be extended in a straightforward way if the seller's goal also concerns the turnover, \emph{i.e.}, by defining $R_t(\boldsymbol{p}_t, \boldsymbol{\omega}_t)$ as a convex combination of turnover and per-round profit.}

However, in a real-world scenario, the functions $\{ v_i(\cdot, \cdot, \cdot) \}_{i=1:\eta}$ are unknown to the seller and, therefore, need to be estimated using the transactions collected over time.
Notice that the volumes $v_i(\boldsymbol{p}_t, \boldsymbol{\omega}_t, t)$ for the $i$-th volume interval also depend on the choices of the other prices and thresholds, as users might be more prone to purchase more items if there is a significant difference in
price than buying fewer products in a single round.
In this way, the problem can naturally be cast as a Multi-Armed Bandit (MAB) problem (see, \emph{e.g.}, \citet{lattimore2020bandit} for a comprehensive review of MAB methods) where the goal is to properly balance the acquisition of information on the functions 
$v_i(\cdot, \cdot, \cdot)$, while maximizing the cumulative reward, a.k.a.~exploration/exploitation dilemma.
Formally, in a MAB problem, we are given a set of available options (a.k.a.~\emph{arms}), and we choose an arm at each time $t$. 
In our case, the arms are all the possible prices $\boldsymbol{p}_t$ and thresholds $\boldsymbol\omega_t$, and the goal is to 
maximize the reward (in our setting profit) over a time horizon of $T$ round.
A policy $\mathfrak{U}$ is an algorithm that returns at each time $t$ a pair $(\boldsymbol{p}_t, \boldsymbol{\omega}_t)$ based on 
the information, \emph{i.e.}, volumes $\tilde{v}_{it}$ and corresponding prices $p_{it}$, we collected in the previous $t-1$ rounds.
A policy is evaluated in terms of \emph{average total reward}, \emph{i.e.}, our goal is to design policies that maximize:
\begin{equation}
	R_T(\mathfrak{U}) := \sum_{t=1}^T \sum_{i=1}^\eta (p_{it} - c) \cdot v_{i}(\boldsymbol{p}_t, \boldsymbol{\omega}_t, t).
\label{eq:reward}
\end{equation}
It is common in the MAB literature to use regret instead of reward as a performance metric. However, the minimization of the former corresponds to the maximization of the latter; therefore, our goal is the standard in MAB settings. Here, the total reward has been selected as a performance metric since it does not require knowledge of the optimum price strategy, which is unknown in the real world.

\section{Algorithm}
\label{sec:algorithm}

The problem presented before is computationally heavy (\emph{i.e.}, exponential in the number of thresholds $\eta$) and cannot be addressed effectively in the presence of scarce data. Indeed, estimating the volume functions $v_{i}(\boldsymbol{p}_t, \boldsymbol{\omega}_t, t)$, each of which has $2 \eta + 1$ input parameters, would take a long time due to the requirement of collecting a large amount of transaction data. In what follows, we approximate the original problem in two different directions to allow learning the volume functions in a short time.

\begin{figure*}[t!]
    \centering
    \resizebox{0.66\linewidth}{!}{
    \includegraphics[]{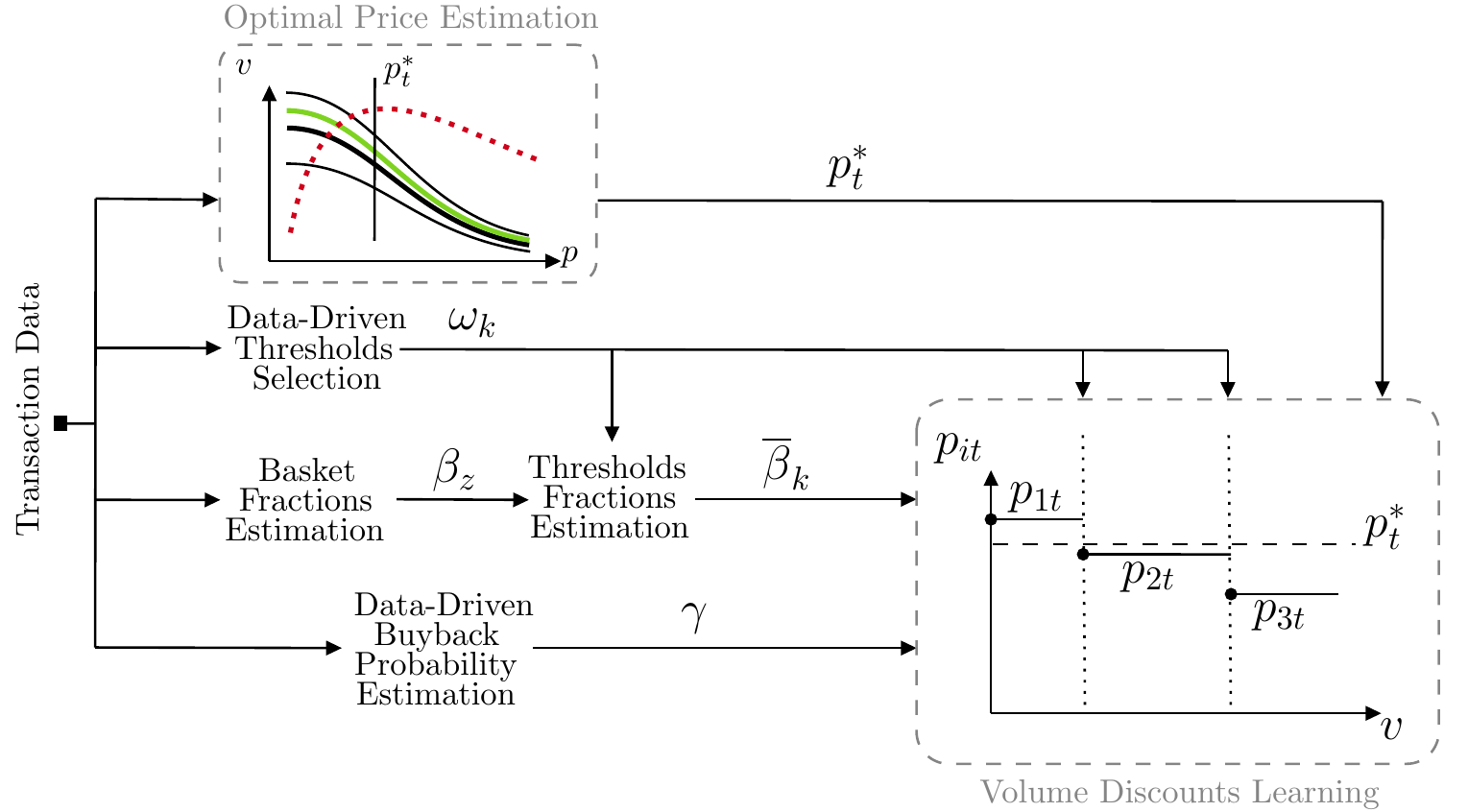}}
    \caption{General overview of the \alg{} algorithm.}
    \label{fig:volume_disc_flow}
\end{figure*}

We assume that the $i$-th volume function depends only on the price $p_{it}$ selected for the $i$-th interval and on time $t$, or, formally, $v_{i}(\boldsymbol{p}_t, \boldsymbol{\omega}_t, t) = v_i({p}_{it}, t)$. Let us define the function of the total volumes provided by a pricing strategy $(\boldsymbol{p}_t, \boldsymbol{\omega}_t)$ as:
\begin{equation} \label{eq:totalvolume}
    \bar{v}(\bar{p}_t, t) := \sum_{i=1}^\eta v_i({p}_{it}, t),
\end{equation}
where $\bar{p}_t$ is a weighted average value of the prices vector $\boldsymbol{p}_t$, formally:
\begin{equation} 
    \bar{p}_t = \sum_{i=1}^{\eta} \alpha_{it} \cdot p_{it},
\label{eq:averageprice}
\end{equation}
where $\alpha_i \in [0,1], \forall i \in \{ 1, \ldots, \eta \}$ must be estimated guaranteeing that on average a threshold pricing strategy $\{\omega_{it}, p_{it}\}_{i=1}^{\eta}$ yields a reward greater than or equal to the theoretical one formulated as follows:
\begin{equation}
    \bar{R}_t(\bar{p}_t) := (\bar{p_{t}} - c) \cdot \bar{v}(\bar{p}_t, t).
    \label{eq:theor_reward}
\end{equation}
Thanks to the previous definitions, we can reformulate the optimization problem into two consecutive steps:
\begin{itemize}
    \item Finding a single optimal price $p^*_t$ that maximizes the revenue provided by the total volume function defined as $R^*_t(p^*_t) := (p^*_{t} - c) \cdot v(p^*_t, t);$
    \item Given $p^*_t$, find a pricing strategy $(\boldsymbol{p}^*_t, \boldsymbol{\omega}^*_t)$ whose weighted average (see Equation~\ref{eq:averageprice}) to the optimal price $p^*_t$.
\end{itemize}
Notice that the second step allows the algorithm to use all the data available for estimating the function $v(\cdot, \cdot)$, instead 
of partitioning them into $\eta$ disjoint sets and independently estimating the $v_i(\cdot, \cdot)$ functions. This allows us to speed up the learning process.
In what follows, we detail the two phases of the \alg algorithm, which corresponds to the solution of the above problem, \emph{i.e.}, the \textit{Optimal Price Estimation} 
and \textit{Volume Discounts Learning} phases. The overall procedure is depicted in Figure~\ref{fig:volume_disc_flow}.
More specifically, the former phase aims to estimate the optimal price $p^*_t$ for the total volumes relying on the transaction data. 
Instead, the latter phase combines the previous estimate of the optimal price $p^*_t$ with the parameters extracted from the transaction data to compute the optimal thresholds $\boldsymbol{\omega}^*_t$ and the pricing strategy $\boldsymbol{p}^*_t$.

\subsubsection{Optimal Price Estimation}
\label{subsec:demandestimation}

As a first step, we estimate the optimal average price $p^*_t$.
The algorithm takes as input the records of past orders, \emph{i.e.}, tuples $(\tilde{p}_{it}, \tilde{v}_{it}, t)$ with the price, 
volume and time of each user purchase occurred in the past time instants.
For each time $t$, it computes the tuple $(\bar{p}_{t}, \bar{v}_t, t)$, where the average price $\bar{p}_{t}$ and the total volume 
$\bar{v}_t$ is computed as described in Equations~\eqref{eq:averageprice} and~\eqref{eq:totalvolume}, respectively, 
relying on the above-mentioned collected data.

\begin{figure*}[t!]
    \centering
    \resizebox{0.66\linewidth}{!}{
    \includegraphics[]{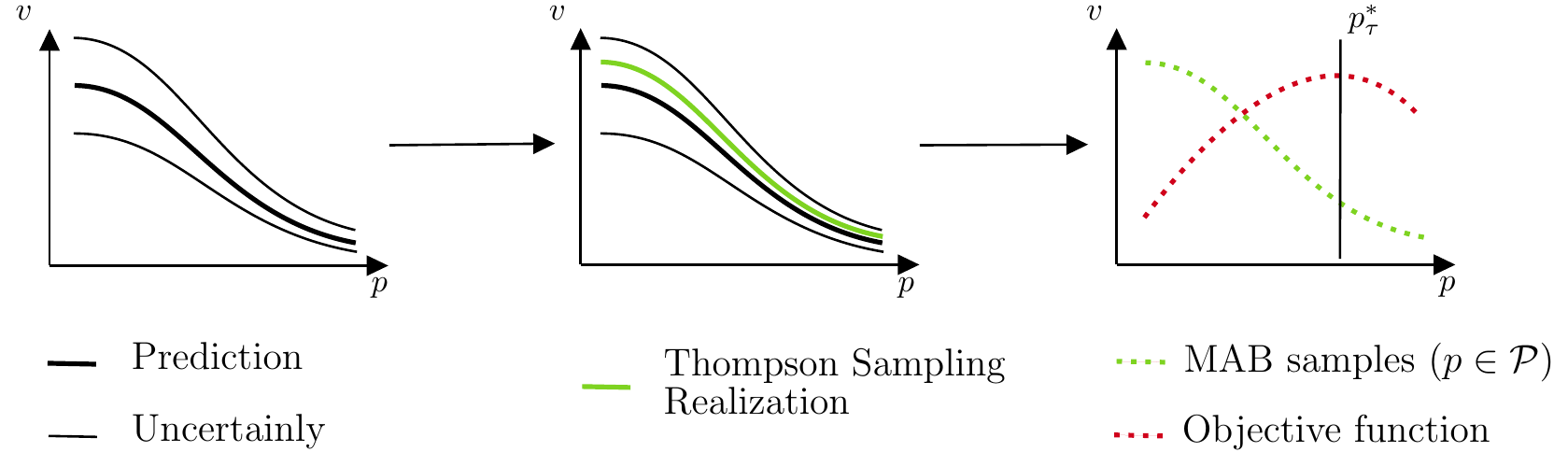}
    }
    \caption{Procedure to retrieve optimal price $p_\tau^*$ at time $\tau$ using a TS-like approach.}
    \label{fig:model_estim_flow}
\end{figure*}

These data are used to compute an estimate $\hat{v}(\cdot, \cdot)$ of the total volume function $v(\cdot, \cdot)$.
The \alg{} algorithm resorts to a Bayesian Linear Regression~\cite[BLR,][]{tipping2001sparse} model to approximate the function $v(\cdot, \cdot)$.
Formally, the estimates of the total volume for price $p$ at time $\tau$ is equal to: 
\begin{equation}
    \hat v(p, \tau) = \sum_{u=1}^U \theta_{u} \phi_{u}(p) + \sum_{d=1}^D \theta_{d} \phi_{d}(\tau),
\end{equation}
where $\phi_1(p), \ldots, \phi_U(p)$ are the basis functions constructed over the price $p$ having as prior a Lognormal distribution and $\phi_1(\tau), \ldots \phi_D(\tau)$ are the basis functions constructed over the time $\tau$ having as prior a Gaussian distribution.
Two remarks are necessary.
First, the basis $\phi_d(\cdot)$ has been introduced to consider the seasonality that affects the selling process of the e-commerce website.
Second, the choice of Lognormal prior for the price basis $\phi_u(\cdot)$ forces such distributions to be non-negative, which in turn induces the monotonicity of the approximating function w.r.t.~the price $p$ (see~\citet{wilson2020bayesian} for more details).
As discussed before, this property is also commonly reflected by real demand functions and allows for fast learning of such curves.

The output of the BLR regression model provides a distribution $\hat v(p, \tau)$ for each $p$, allowing the use of MAB algorithms and, in particular, the use of a Thompson Sampling (TS)-like~\cite{kaufmann2012thompson,agrawal2012analysis} approach, as a strategy to find a value for the optimal price $p^*_\tau$ balancing exploration and exploitation. The corresponding procedure is summarized in Figure~\ref{fig:model_estim_flow}.
More specifically, from the distribution $\hat v(\cdot, \tau)$ (Left, represented as expected values and uncertainty bounds), we sample a function $\hat{v}_{TS}(\cdot, \tau)$ (Center, represented in green), and, finally, we perform the optimization (Right) of the profit as follows:
\begin{equation}
    p_\tau^* \in \argmax{p \in \mathcal{P}} (p-c) \cdot \hat v_{TS}(p, \tau).
\end{equation}

\subsubsection{Volume Discounts Learning}
\label{subsec:volumediscounts}

Let $\eta$ be the number of volume thresholds to show to customers, along with the corresponding prices.\footnote{Here, we assume the e-commerce experts provide this value. 
If not provided, clustering techniques over historical data can be used to determine the optimal number of groups in the user distribution, \emph{e.g.}, \citet{pelleg2000x}.}
Let $\beta_{z}$, with $z \in \mathbb{N}$, be the proportion of baskets containing the product with a volume of $z$. 
The average volume for the product in each basket is $\bar{V} = \sum_{i=1}^{\infty}\beta_{i}\cdot i$. 

Given the threshold $\omega_{k}$, the total proportion of baskets inside that range is given by:
\begin{equation}
    \bar{\beta}_{k}
    =\sum_{i=\omega_{k}}^{\omega_{k+1}-1}\beta_{i}.
\end{equation}
The average volume of products for the baskets in a given threshold $\omega_k$ is consequently defined as:
\begin{equation}
    \bar{V}_{k} = \frac{\sum_{i=\omega_{k}}^{\omega_{k+1}-1} \beta_{i} \cdot i}{\sum_{i=\omega_{k}}^{\omega_{k+1}-1} \beta_{i}}.
\end{equation}

Suppose a customer needs $N$ units of the given product.
This need can be fulfilled by dividing his/her order across any number of time steps, \emph{i.e.}, performing a purchase of units (or a \textit{volume}) 
in a range $\{\omega_{k},\ldots, \omega_{k+1}-1\}$ for a specific $k$ and repeating such a purchase until the required amount of units is reached.
After the customer bought the product, he/she has a probability $\gamma$ of returning to the same retailer 
buying another batch of the same size. This kind of modeling of the user's behavior is reflecting accurately those customers buying goods 
with a short lifespan and, in general, the ones for which a customer is led to schedule periodic purchases (\emph{i.e.}, toilet paper, consumable office supplies).
With probability $1 - \gamma$, the customer will not return the next time. We consider $\gamma$ a property of the system, 
so we assume that the price does not affect the buyback probability.
In what follows, we use historical transaction data to estimate the value of $\gamma$ for a given product.

Let $\bar m$ denote the desired margin when a single unit is purchased. It follows that the expected margin $\bar \mu$ coming 
from a customer with a need of $N$ units and who performs only single-unit orders is:
\begin{equation}
\label{eq:rew_single_unit}
    \bar \mu = \sum_{\tau=1}^{N} \gamma^{\tau-1}\bar m = \frac{1-\gamma^N}{1-\gamma} \bar m,
\end{equation}
where the last equality comes from the truncated geometric series identity.
A customer with the same need, but whose orders contain a number of units in $\{\omega_{k},\ldots, \omega_{k+1}-1\}$, which are associated with a margin $\bar m_{k}$, will generate the following expected margin:
\begin{equation}
\label{eq:rew_k_unit}
\bar \mu_{k} = \sum_{\tau=1}^{\left\lceil \frac{N}{\bar{V}_{k}}\right\rceil} \gamma^{\tau-1} \bar m_k \bar{V}_{k} = \frac{1-\gamma^{\left\lceil \frac{N}{\bar{V}_{k}}\right\rceil}}{1-\gamma}(1-\delta_k) \bar m \, \bar{V}_{k},
\end{equation}
where $\delta_k$ is the discount applied to the single-unit margin $\bar m$, namely:
\begin{equation}
    \label{eq:margin_disc}
    \bar{m}_k = \bar{m}(1-\delta_k), \hspace{1cm} k \in \{1, \ldots, \eta\},
\end{equation}
where $\delta_1=0$.
By imposing $\bar\mu_k \geq \bar\mu_1$, we get:

\begin{equation}\label{eq:discount}
    \delta_k \leq 1 - \frac{1-\gamma^N}{\bar{V}_{k}\left(1-\gamma^{\left\lceil \frac{N}{\bar{V}_{k}}\right\rceil}\right)}.
\end{equation}

Given the desired margin $m^*_t=p^*_t-c$ derived in the previous section, the expected profit without any discount can be computed as $m_t^*\bar V$. 
Suppose we are applying a volume discount policy: we expect it will not decrease the total expected margin given without it. 
Unit-volume margin $\bar m$ can be computed by imposing that the expected margin without any discount policy coincides with the one including them:
\begin{equation}
    \label{eq:margin_identity}
    \sum_{k=1}^\eta \bar V_k\bar m_k = m_t^* \bar V.
\end{equation}
Substituting Eq.~\ref{eq:margin_disc} into Eq.~\ref{eq:margin_identity}, we get:
\begin{equation}
    \label{eq:vol_discount}
    \bar m = \frac{m_t^* \bar V}{\sum_{k=1}^\eta (1-\delta_k)\bar V_k}.
\end{equation}
Finally, the margins $\bar m_1, \ldots, \bar m_\eta$ for the different volume thresholds are determined as $\bar m_k = \bar m (1-\delta_k)$, for $k \in \{1, \ldots, \eta\}$, where $\delta_1 = 0$.
The complete algorithm, including both optimal average price estimation and volume discounts, is summarized in Figure~\ref{fig:volume_disc_flow}.

\subsubsection{Data-Driven Buyback Probability Estimation}
Even if estimating $\gamma$ in an online fashion would be a natural approach, it is prohibitive due to our environment's strong seasonality and non-stationary nature. 
Indeed, studying customers' churn usually requires a large amount of contextual data and is a challenging task for many fields~\citep{kamalraj2013survey}.
Instead, we propose a methodology purely based on the available transaction data (where the customers are uniquely identified) to estimate $\gamma$ in an offline fashion. 
We define two time-intervals: a \textit{``measure"} period $\mathcal{T}_M \in \bar{\mathcal{T}}$ and a \textit{``control"} one $\mathcal{T}_C \in \bar{\mathcal{T}}$, where $\mathcal{T}_M \cap \mathcal{T}_C = \emptyset$ and $\bar{\mathcal{T}}$ is the set of time periods \emph{i.e.}, contiguous sequences of times.
Intuitively, we observe which customers buy during the \textit{``measure"} period and compute what percentage of them come back in the subsequent period, the \textit{``control"} one.
Formally, given a set of customers $\mathcal{G} := \{ g_1, \ldots, g_L\}$, we define a function $\mathcal{H} : \mathcal{G} \times \bar{\mathcal{T}} \rightarrow{} \mathbb{N}$ that associates a customer to the number of purchases made in a specific period. 
We also introduce $h : \bar{\mathcal{T}} \rightarrow{} \mathcal{P}(\mathcal{G})$, which maps a period of time into the subset of unique customers who made at least one purchase in that period.\footnote{With $\mathcal{P}(A)$ we denote the power set of $A$.}
Thus, we are able to compute the total number of returns $\mathcal{R}$ and non-returning customers $\mathcal{A}$, formally $\mathcal{R} = \sum_{g\in\mathcal{G}} \left[\mathcal{H}(g,\mathcal{T}_M)\right] - |h(\mathcal{T}_M)| + |h(\mathcal{T}_M) \cap h(\mathcal{T}_C)|,$ and $\mathcal{A} = |h(\mathcal{T_M})|-|h(\mathcal{T}_M) \cap h(\mathcal{T}_C)|$. Notice that $\mathcal{R}$ is composed by those customers that have already occurred during the \textit{``measure"} 
period (since a customer that purchased $n$ times during $\mathcal{T}_M$ already returned $n-1$ times) and those happened during 
the \textit{``control"} period (customers seen in both periods).
Instead, $\mathcal{A}$ is the number of customers who purchased at 
least one time in the \textit{``measure"} period and did not show up during the \textit{``control"} one. Notice that the functions 
$\mathcal{H}$ and $h$ can be easily calculated starting from transaction data once the two periods have been defined. 
In our test, we decided to use the $6$ months before the experimental campaign as a \textit{``control"} period and the previous $6$ months as a \textit{``measure"} one.
Finally, thanks to the two quantities defined above, the value of $\gamma$ can be approximated as $\gamma = \frac{\mathcal{R}}{\mathcal{R}+\mathcal{A}}$.

\begin{figure}[t!]
    \centering
    \resizebox{0.835\columnwidth}{!}{\includegraphics[]{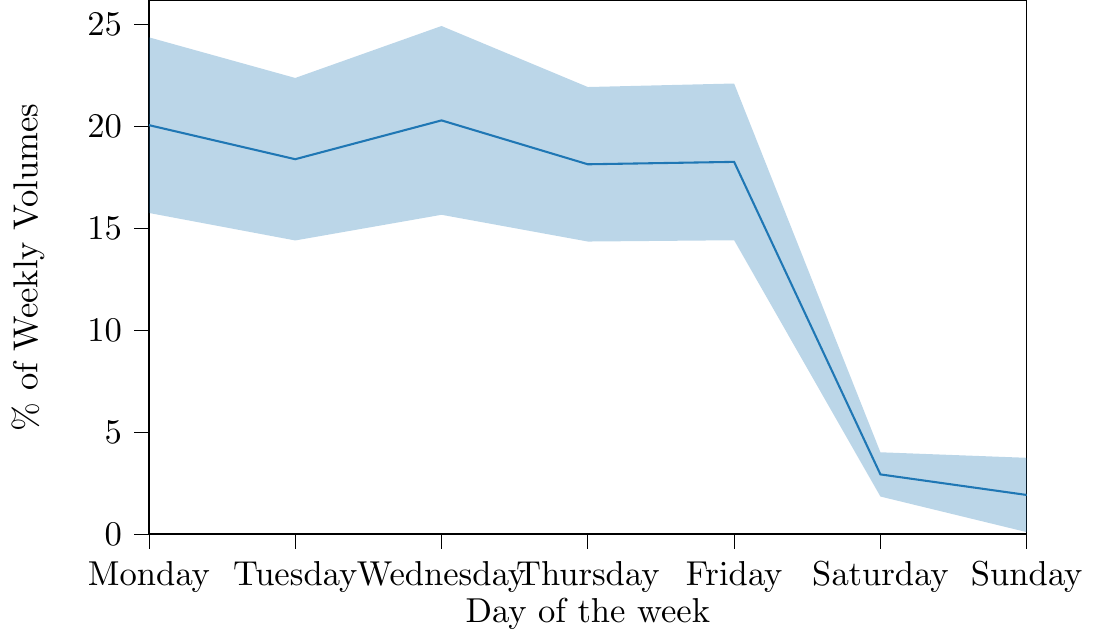}}
    \caption{Seasonality over a single week (mean $\pm$ std).}
    \label{fig:intraweekseasonality}
\end{figure}

\subsubsection{Data-Driven Threshold Selection}
Threshold values $\{ \omega_k \}_{k=1}^{\eta}$ can be selected in several ways. 
Our solution is to define a split criterion that divides the products within the shopping baskets into $\eta$ sets of equal cardinality.
Formally, we define $q : \mathbb{N} \rightarrow{} \mathbb{N}$ as the function that maps a number of units into the number of shopping baskets that contain that many units.
Notice that if $B$ is the total number of shopping baskets over the period examined, then $q(z) = B \cdot \beta_z$, where the values of $\beta_z$ and $B$ can be estimated from the transaction data, and, consequently, $q(\cdot)$ can be computed entirely from data.
Intuitively, we can build a data set where each $z \in \mathbb{N}$ is repeated $q(z)\cdot z$ times, defined as:
\begin{equation}
    Q := \{\underbrace{1,\ldots,1}_{\textrm{$q(1)$ times}},
    \ldots, \underbrace{d,\ldots,d}_{\textrm{$d\cdot q(d)$ times}},\ldots\}.
\label{eq:thresh_dataset}
\end{equation}
To get the $k$-th threshold, we extract the $\left \lceil |Q|\cdot \frac{k}{\eta} \right \rceil$-th element from the sorted data set $Q$ defined as in Eq.~\ref{eq:thresh_dataset}.

\section{Experimental Evaluation}
\label{sec:experiments}

We performed a real-world experiment in collaboration with an Italian e-commerce company in which our algorithms priced a set of products adopting a long-tail economic model~\cite{anderson2006long}.
The e-commerce website collects data on each purchase (date and time), as a row of a transaction data set, including features such as the identifier of the purchased product, the number of units sold, the price, the cost, and the class of the customer (business or private) inferred from the fiscal status observed after the purchase.
The experimental campaign focused on products usually bought with high volumes to evaluate our algorithms better.

\begin{figure}[t!]
    \centering
    \resizebox{0.77\columnwidth}{!}{\includegraphics[]{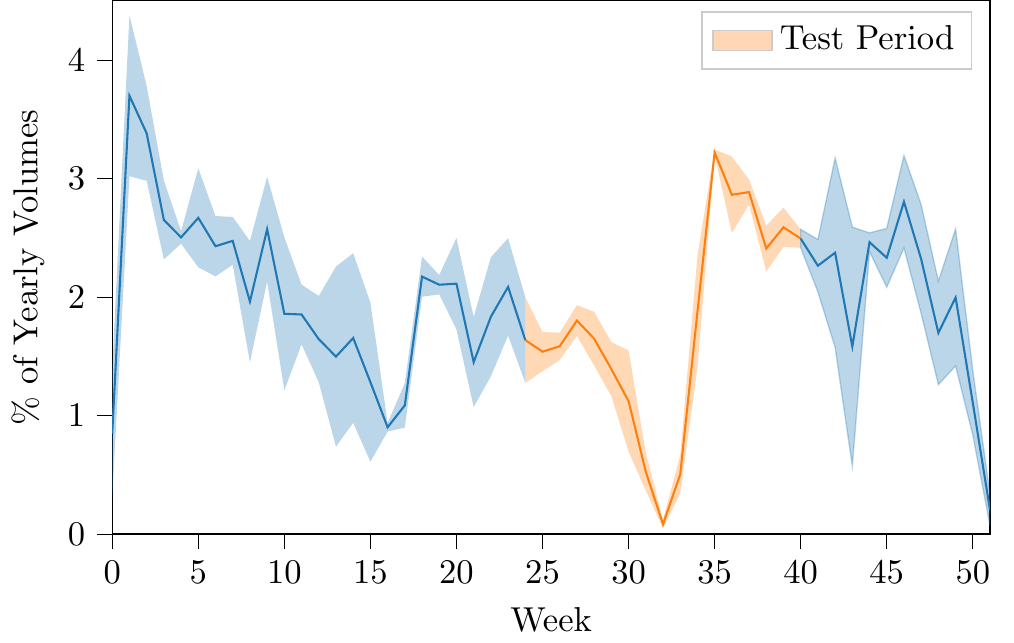}}
    \caption{Seasonality over the weeks of a year (mean $\pm$ std).}
    \label{fig:seasonality}
\end{figure}

We resorted to an online \textsf{A}/\textsf{B}.
The experimental campaign was conducted in one of the main categories of the e-commerce website, with a test set (\textsf{A}) composed of $N_t = 295$ products and a control set (\textsf{B}) composed of $N_c = 33$ 
products of the same category and with the same characteristics.\footnote{
The test and the control sets were defined by e-commerce specialists according to both technical and market issues.}
The test included products with a yearly turnover of $300$ KEuros and a total profit of $83$ KEuros. 

The algorithm produces new prices every $7$ days since a significant intra-week seasonality has been observed (see Figure~\ref{fig:intraweekseasonality}). 
Moreover, the products sold by the e-commerce website are subject to a significant seasonality over different periods of the years, as shown in Figure~\ref{fig:seasonality}.
Due to the particular kind of products sold we dealt with and the nature of the target customer segment, volume discounts are crucial to the business 
since they affect customers' loyalty and the logistic organization of the company. The e-commerce website's specialists defined the number $\eta = 3$ of volume thresholds that should be displayed for every product.
The test was conducted for $17$ weeks, from $16$ June $2021$ to $17$ October $2021$, during which no communication and marketing actions were performed in attempt not to influence the customers' behavior.

\begin{figure}[t!]
    \centering
    \resizebox{\columnwidth}{!}{\includegraphics[]{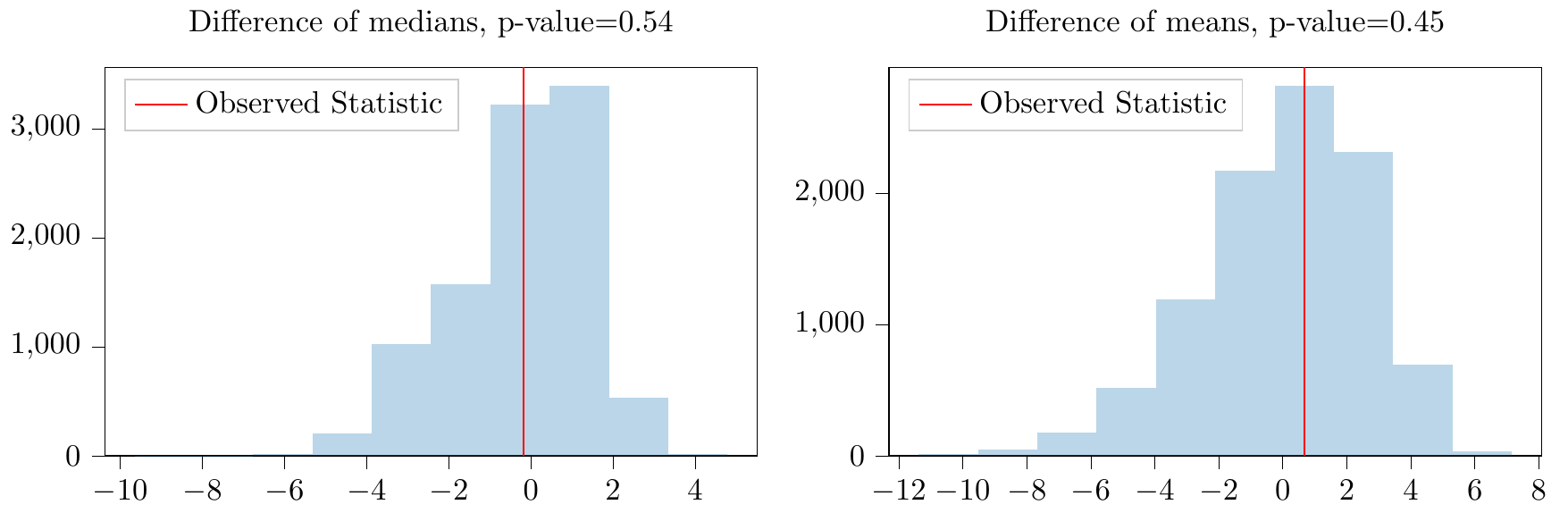}}
    \caption{Distribution of the two-sided permutation tests statistics before the test, $R=10000$ random permutations.}
    \label{fig:priori_perm_test}
\end{figure}

\begin{figure}[t!]
    \centering
    \resizebox{\columnwidth}{!}{\includegraphics[]{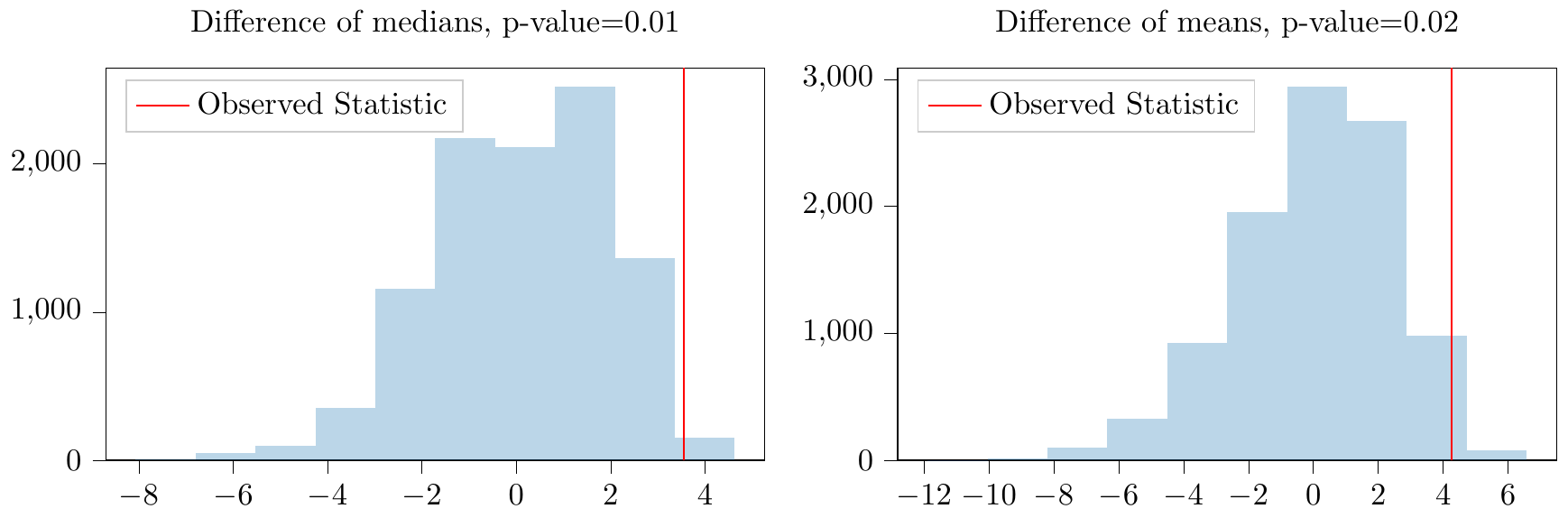}}
    \caption{Distribution of the two-sided permutation tests statistics after the test, $R=10000$ random permutations.}
    \label{fig:posteriori_perm_test}
\end{figure}

The business goal was to maximize test set's (\textsf{A}) average profit $R^{(\text{\textsf{A}})}_T$, as defined in Eq.~\eqref{eq:reward}, where $T = 17$.
This score is to be compared with that one achieved by the \textsf{B} set $R^{(\text{\textsf{B}})}_T$ over the same period. To evaluate the performance of our algorithm, we performed a statistical test applied to the product margins to check whether the two groups are comparable.
More specifically, for each product, we computed the average weekly net margins ($R_t$ as in Eq.~\ref{eq:inst_reward}) during the first six months of $2021$ (\emph{i.e.}, $t$ is in the first $26$ weeks of $2021$) and we design a test to check if the median and mean of the net margin across the products of the \textsf{A} set are larger than the ones of the \textsf{B} set.
We performed one-sided permutation tests with the null hypothesis being ``The \textsf{A} set has not a higher median/mean of net margin w.r.t.~the 
\textsf{B} set". Figure~\ref{fig:priori_perm_test} shows the distributions of the tests' statistics together with the observed one, in which the 
resulting p-values concerning medians and means are respectively $0.54$ and $0.45$, and, therefore, resulting in the fact that there is not enough statistical evidence to say the two are different.
This shows that set \textsf{A} has not a larger median/mean w.r.t.~set \textsf{B} on the chosen performance metric before the beginning of the test.

We use the BLR basis functions following different criteria to perform the test mentioned above. To model the price elasticity over 
the customers' base, we choose \textit{reverted hyperbolic tangent} functions. Instead, to grasp the irregular nature of e-commerce's seasonality, we choose \textit{Radial Basis Functions}~(RBF). Finally, the trend is modeled by choosing polynomial basis functions. 
Both RBF and reverted hyperbolic tangents are evaluated with different shifts and scales, while polynomial features with different degrees.

The algorithm ran in a Docker container with \textit{Python 3.8} environment on \textit{Linux}. Every week, the algorithm made an SQL query to retrieve that data about the products and then returned the prices. The hardware was a \textit{Quad-core Intel Core i7 8th Gen} with \textit{8Gb DDR4 RAM}. The time required for a run of the algorithm over all the products is about 25 minutes. Given that the algorithm is applied to each product independently of the others, the running time scale linearly w.r.t.~the number of products.

\subsubsection{Results}

The goods priced by \alg{} during the testing period provided an improvement (on average) in terms of the performance metric $R^{(\text{\textsf{A}})}_T$ of $+$\perf{} w.r.t.~the one $R^{(\text{\textsf{B}})}_T$ of control set of goods, or formally $\frac{R^{(\text{\textsf{A}})}}{R^{(\text{\textsf{B}})}} = 1.55$.
After $17$ weeks, we performed the same statistical test on the weekly performance metric obtained between the two sets of products during the test period.
Figure~\ref{fig:posteriori_perm_test} shows the distribution of the test's statistics along with the observed ones.
The two tests, performed with the same seed and number of random permutations of the previous, yielded this time $p$-values on the medians and the means of respectively of $0.01$ and $0.02$, allowing us to reject the null hypothesis and conclude that the test set of products has both a larger median and mean of the average weekly performance metric w.r.t.~the control set of products, with at least a confidence of $98\%$.

\begin{figure}[t]
    \centering
    \resizebox{0.8\columnwidth}{!}{\includegraphics[]{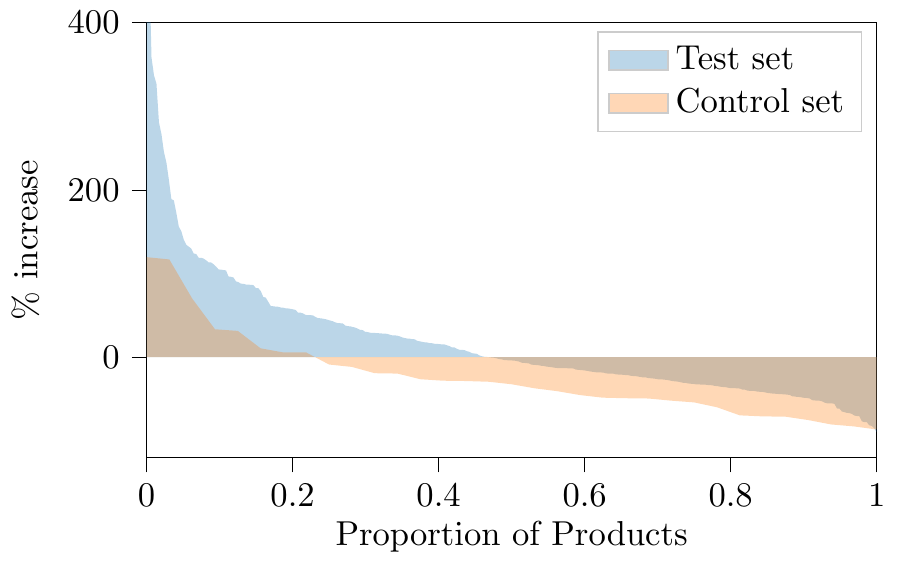}}
    \caption{Distribution of the objective function improvements in Test and Control set.}
    \label{fig:margin_improv}
\end{figure}

Regarding the performances on a product-wise level, we report in Figure~\ref{fig:margin_improv} the sorted percentages of improvement 
on the performance metric w.r.t.~to the period of $2021$ preceding the test for every single product.
In the test set, $138$ products over $295$ ($ \approx 47\% $) improved their average performance w.r.t.~the 
corresponding period of $2021$, while in control set only $8$ products over $33$ ($ \approx 25\% $) were able to improve.
This corroborates the idea that the proposed method is able to improve the performance of the e-commerce website by influencing the purchase process of a large number of products.

\subsection{Effect of Volume Discounts}

A final analysis consists in evaluating how the volume discounts algorithm can modify the probability distribution of the units count of the same product in a basket. 
More precisely, we need to check whether the algorithm affects the volumes $\bar{\beta}_k$ to increase the profit.
In our specific setting, this corresponds to checking if the value $\bar{\beta}_1$ decreases in favor of $\bar{\beta}_2$ and/or $\bar{\beta}_3$.
For this analysis, the parameters of the volume-discount algorithm have been estimated using the period from $16$ June $2019$ to $16$ June $2021$, and the estimation of $\gamma$ was performed on the data split in an estimation period $\mathcal{T}_M$ from $17$ June $2019$ to $16$ June $2020$ and a control one $\mathcal{T}_C$ from $17$ June $2020$ to $16$ June $2021$.

\begin{figure}[t]
    \centering
    \resizebox{0.8\columnwidth}{!}{\includegraphics[]{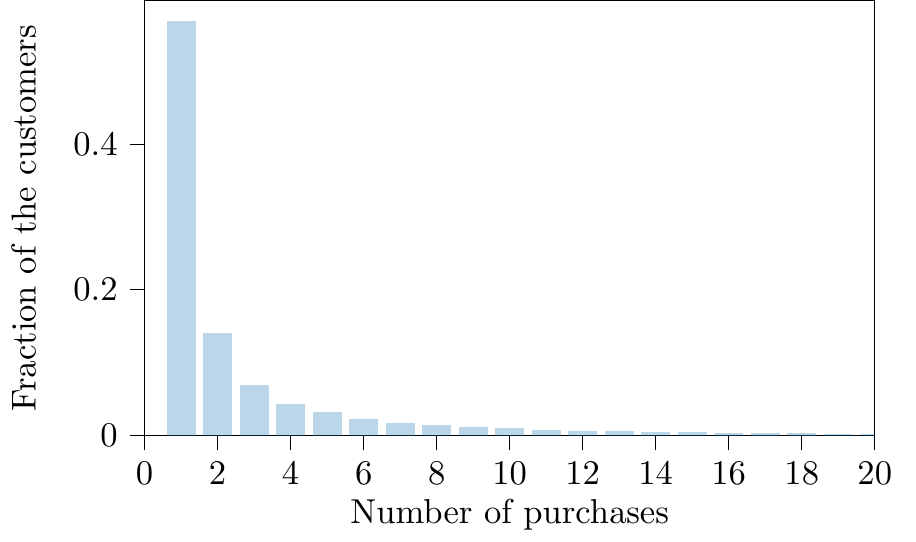}}
    \caption{Number of purchases done by the customers.}
    \label{fig:purchases_distr}
\end{figure}

\begin{figure}[t]
    \centering
    \resizebox{0.8\columnwidth}{!}{\includegraphics[]{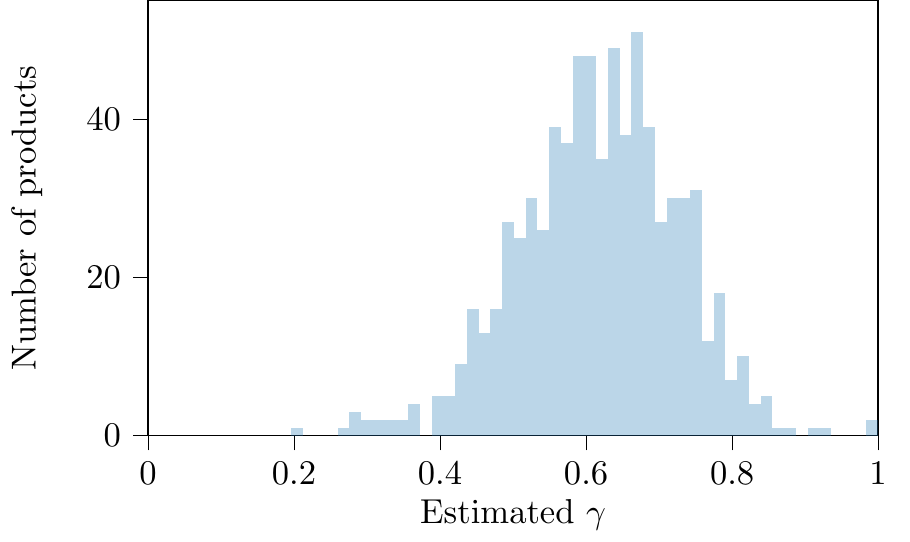}}
    \caption{Distribution of the parameter $\gamma$.}
    \label{fig:gamma_distr}
\end{figure}

\begin{figure}[t]
    \centering
    \subfloat[From $p_{1t}$ to $p_{2t}$.]{\resizebox{0.533\linewidth}{!}{\includegraphics[]{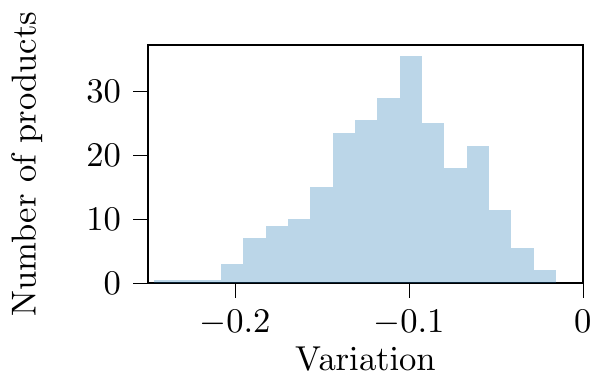}} \label{fig:p1_p2}}
    \subfloat[From $p_{1t}$ to $p_{3t}$.]{\resizebox{0.47\linewidth}{!}{\includegraphics[]{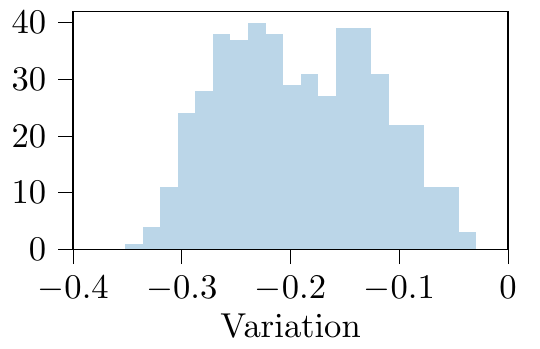}} \label{fig:p1_p3}}
    \caption{Average (on time) discounts between volumes' thresholds in test products.}
    \label{fig:vol_disc_price_vars}
\end{figure}

We analyze the context from both customers' and products' perspectives.
Figure~\ref{fig:purchases_distr} shows the distribution of the number of orders performed by the customers. The histogram shows almost half of the customers perform a single order and leave the shop without coming back. Figure~\ref{fig:gamma_distr} analyze the same phenomenon from the product perspective. Given a product, we can interpret the $\gamma$ parameter as the probability that a customer buying that product will, sooner or later, buy such product again. Higher is the $\gamma$, the more conservative will be the discount strategy we adopt.

\subsubsection{Results}
The estimated per-product discounts between the three thresholds are presented in Figure~\ref{fig:vol_disc_price_vars}.
The \alg{} algorithm applies an average discount of $\approx 10\%$ for the second volume interval and $\approx 20 \%$ for the third one.
This implies a shift in the number of purchases among the intervals.
In Table~\ref{tab:delta_beta}, the variations of the three $\bar{\beta_k}$ are reported: during the test, we achieved an increase of the values $\bar{\beta_2}$ and $\bar{\beta_3}$ while observing a reduction in $\bar{\beta_1}$.
Finally, in Table~\ref{tab:delta_beta_2} we report the average variations in terms of average units per basket of the $4$ above-mentioned products during the test period.
The effect of applying the \alg{} algorithm and, therefore, introducing volume discounts modifies the basket's average size by increasing the units purchased by $\approx 33 \%$.

\subsection{Considerations After the A/B Test}
After the end of the \textsf{A}/\textsf{B} test, the e-commerce specialists were satisfied with the achieved results, including the performance of the volume-discount algorithm.
Thus, the company extended the adoption of our algorithm to all the products presenting a sufficient amount of volumes in the catalog of the e-commerce website ($\approx 1200$ products). Currently, our algorithm prices about $1,200$ products generating a cumulative annual revenue of about $1.5$ MEuro, which corresponds to about  $50\%$ of the total e-commerce website turnover. Furthermore, the algorithm now runs in the cloud in a SaaS fashion. An automatized routine runs a query on the dataset of the e-commerce website, extracting the transaction data needed and then running the algorithm. The results are provided to the business unit in a \texttt{csv} file.

\begin{table}[t!]
    \centering
    \caption{Variations of $\bar{\beta}_k$ after the test period.}
    \label{tab:delta_beta}
    \small
      \begin{tabular}{cccc}
          \toprule
          Product \; & $\Delta \bar{\beta_1}$ & $\Delta \bar{\beta_2}$  & $\Delta \bar{\beta_3}$  \\
          \midrule
          \midrule
          1 &	-32\% & +10\% & +22\% \\
          2 &	-26\% &	+25\% &	+1\% \\
          3 &	-15\% &	+4\% & +11\% \\
          4 &	-5\% & +1\% & +4\% \\
          \midrule
          Mean & -19.5\% & +10\% & +9.5\% \\
          \bottomrule
      \end{tabular}
  \end{table}

\begin{table}[t!]
    \centering
    \caption{Variation of units per basket after the test period.}
    \label{tab:delta_beta_2}
    \small
      \begin{tabular}{cc}
          \toprule
          Product \; & $\Delta$units \\
          \midrule
          \midrule
          1 & +63\% \\
          2 & +43\% \\
          3 & +11\% \\
          4 & +14\% \\
          \midrule
          Mean & +33\% \\
          \bottomrule
      \end{tabular}
  \end{table}

\section{Conclusion and Future Works}
\label{sec:conclusion}

In this paper, we present \alg{}, an algorithm capable of defining the price and volume discounts in an online setting. 
Our approach exploits the transaction data of the e-commerce website to optimize the pricing strategy in an online fashion.
We test our approach in a real-world $4$-months experiment by optimizing the price of $295$ products of an e-commerce website. 
The results show that our approach increases the e-commerce website profits by outperforming the previous management and gaining an increase of \perf{}.

In future works, we plan to insert in the model time correlations between the purchases and the effect of loyalty in increasing revenue. Furthermore, in this work, we price products independently, while cross-selling approaches could further increase profits for some classes of products. The design of algorithms taking into account also these dependencies constitutes an interesting new line of work.

\FloatBarrier

\bibliography{biblio}

\end{document}